\documentclass[sigconf]{acmart}
\usepackage{algorithm,algorithmic}
\usepackage{subfigure}
\usepackage{graphicx}
\usepackage{multirow}
\usepackage{enumerate}
\usepackage{balance}

\AtBeginDocument{%
  \providecommand\BibTeX{{%
    \normalfont B\kern-0.5em{\scshape i\kern-0.25em b}\kern-0.8em\TeX}}}

\setcopyright{acmcopyright}
\copyrightyear{2020} 
\acmYear{2020} 
\setcopyright{acmcopyright}
\acmConference[KDD '20] {26th ACM SIGKDD Conference on Knowledge Discovery and Data Mining}{August 23--27, 2020}{Virtual Event, USA}
\acmBooktitle{26th ACM SIGKDD Conference on Knowledge Discovery and Data Mining (KDD '20), August 23--27, 2020, Virtual Event, USA}
\acmPrice{15.00}
\acmDOI{10.1145/XXXXXX.XXXXXX}
\acmISBN{978-1-4503-7998-4/20/08} 

\allowdisplaybreaks

\begin{document}
\fancyhead{}

\title{Learning Transferrable Parameters for Long-tailed Sequential User Behavior Modeling}

\author{Jianwen Yin $^{1,4,*}$, Chenghao Liu$^{2,*}$, Weiqing Wang$^{3}$, Jianling Sun$^{1,4}$, Steven C.H. Hoi$^{2,5}$}
\affiliation{$^1$ Zhejiang University, $^2$Singapore Management University,$^3$ Monash University}
\affiliation{$^4$Alibaba-Zhejiang University Joint Institute of Frontier Technologies, $^5$ Salesforce Research Asia}
\email{yinjw@zju.edu.cn, chliu@smu.edu.sg, Teresa.Wang@monash.edu, sunjl@zju.edu.cn, shoi@salesforce.com}





\thanks{$^*$ denotes equal contribution.\\
Chenghao Liu and Jianling Sun are corresponding authors. Steven C.H. Hoi is currently with Salesforce Research Asia and on leave from Singapore Management University.}
\begin{abstract}
Sequential user behavior modeling plays a crucial role in online user-oriented services, such as product purchasing, news feed consumption, and online advertising. The performance of sequential modeling heavily depends on the scale and quality of historical behaviors. However, the number of user behaviors inherently follows a long-tailed distribution, which has been seldom explored. In this work, we argue that focusing on tail users could bring more benefits and address the long tails issue by learning transferrable parameters from both optimization and feature perspectives. Specifically, we propose a gradient alignment optimizer and adopt an adversarial training scheme to facilitate knowledge transfer from the head to the tail. Such methods can also deal with the cold-start problem of new users. Moreover, it could be directly adaptive to various well-established sequential models. Extensive experiments on four real-world datasets verify the superiority of our framework compared with the state-of-the-art baselines.
\end{abstract}



\maketitle

\section{Introduction}
With the rapid development of the Internet, the applications of user sequential scenarios have become essential and pervasive, such as e-commerce system, news/articles suggestion, and click-through rate (CTR) prediction \cite{liu2016online,liu2017collaborative,wang2017st,wang2018tpm,yin2019online}. In such applications, each user's behaviors can be represented as sequences in chronological orders, and his/her future behaviors can be predicted with given historical behaviors. Modeling users' complex sequential behaviors is challenging and critically important for providing a personalized recommendation in real-world applications \cite{wang2018attention}. 

\begin{figure}[!htb]
    \centering
    \includegraphics[width=\linewidth]{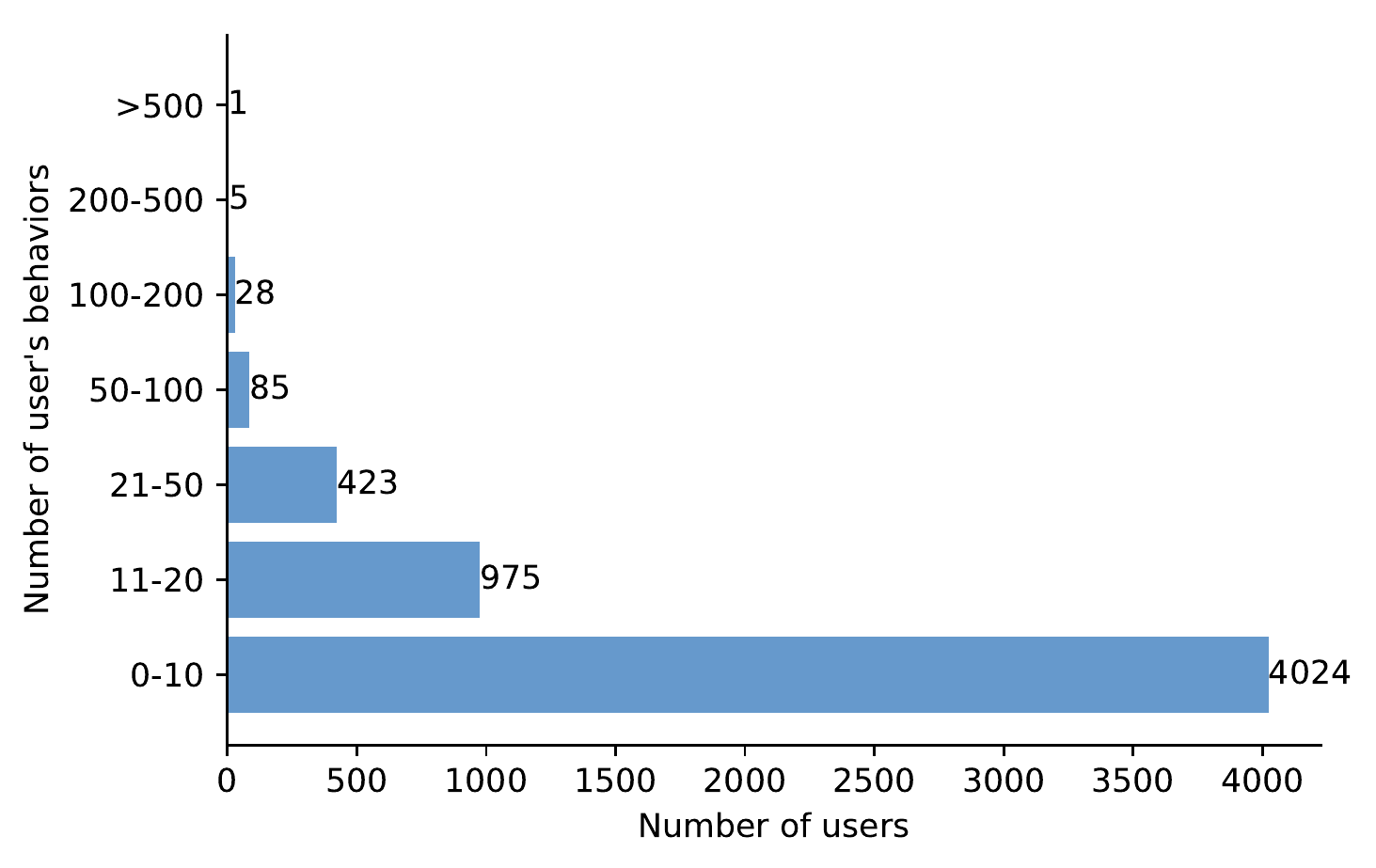}
    \caption{Histogram of the number of users over the number of user's behaviors of the Amazon Music dataset.}
    \label{fig:long-tailed}
\end{figure}

Generally, the performance of sequential user behavior modeling heavily depends on the scale and quality of historical behaviors. To achieve the desired performance, it is required to have enough historical behaviors for each user for sufficient modeling. In real-world applications, however, the number of user's behaviors inherently follows a long-tailed distribution as shown in \ref{fig:long-tailed}, in which the number of historical behaviors of per user varies significantly from hundreds or thousands for head users to as few as one for tail users \cite{beutel2017beyond,sharma2019adaptive}. Although existing approaches for sequential user behavior modeling achieve promising results on those few data-rich head users, they leave many data-poor tail users ill-served. In fact, this common property of real-world datasets encourages a skewed user behavior prediction policy where many tail users are modeled far worse than others.

In contrast, focusing on tail users can bring more benefits: (1) Although head users take a large share of total user-item interactions, the number of tail users is much larger than that of head users. Improving the performance of tail users can significantly increase the retention rate and scale of users, thereby producing massive profits. (2) Compared with the performance over head users where is good enough and has limited room for improvement, tail users embrace relatively large improvement, which means the endeavor to explore the tail users can bring much more improvements. (3) Attention on the tail users can also boost the performance on the head. The infrequent patterns lie in the tail could provide a complementary understanding of the head. (4) In practice, when a learned model is deployed in an online setting, it requires to deal with new users who have no interactions available during training phase \cite{he2016fast}. Focusing on those more diverse and noisy tail users is helpful for learning a robust model that can generalize well on new users.

Given these issues with conventional sequential models, how can we learn a model that is focusing on improving recommendation accuracy for badly-modeled tail users? While it focuses on the tail, could we enhance the general performance of both head users and tail users? Could it improve the performance of new users simultaneously?


In this paper, we aim to address the long-tailed sequential user behavior modeling by learning transferrable parameters. The key idea is that some internal information is more transferrable across users, while others can cause interference. Encouraging the expression of such information during training can improve the performance of tail users, thereby generating an unbiased model which can perform well on both long and tail users. Such a model can also deal with the cold-start problem of new users that are common in real-world applications. Specifically, we address the problem of learning transferrable parameters from both the optimization perspective and feature perspective. In \textbf{optimization perspective}, we develop a gradient alignment optimization method, which could maximize transfer while minimizing interference across users. In the learning process, we update a mini-batch of users with gradient descent. For each pair of users, the gradient angle implies the transferrable ability between these two users. The more consistent the directions of the gradients, the more knowledge they can share. Based on this, we impose an auxiliary loss to maximize the dot product between the gradients generated by different users. In \textbf{ feature perspective}, we introduce a discriminator which takes a sequential embedding as input and classifies whether it belongs to a head user or a tail user. As the model achieves equilibrium, the discriminator cannot well differentiate head users from tail users. Consequently, head users are mixed with tail users in the embedding space. In this way, the bias caused by a large amount of data from head users could be reduced, thus facilitating knowledge transfer from the head to the tail.

The main contributions of this work are summarized as follows:
\begin{itemize}
    \item To the best of our knowledge, this is the first work that addresses the long-tailed sequential user behavior modeling. We argue that focusing on tail users can bring more benefits and achieve this by learning transferrable parameters.
    \item To learn transferrable parameters, we propose a gradient alignment optimizer to transfer knowledge across users from the optimization perspective. Moreover, we introduce an adversarial training method to learn frequency-agnostic sequential embedding, which facilitates knowledge transfer from the feature perspective. The proposed method could be adaptive to various well-established sequential models, such as GRU4REC \cite{hidasi2015session}, CASER \cite{tang2018personalized} and SASR\cite{kang2018self}.
    \item We conduct extensive experiments by evaluating the proposed method on real-world datasets, and show that it outperforms the existing state-of-the-art baselines for sequential user behavior modeling task.
\end{itemize}

\section{RELATED WORK}
In this section, we will review several lines of works closely related to this paper, including sequential user behavior modeling, gradient alignment, and adversarial training.

\subsection{Sequential User Behavior Modeling}
User behavior modeling, which captures users' preferences from behavior data, is critically important since it contributes significant improvement for real-world applications. Researchers have proposed various approaches, from traditional collaborative filtering models \cite{mnih2008probabilistic,koren2009matrix} to deep representation learning models \cite{qu2016product,zhou2018deep}. These models focus on mining the static relationships between users and items, ignoring the dynamics of users' preferences implied in sequential interactions. 

Nowadays, sequential user behavior modeling has attracted considerable attention due to its superiority in capturing item-to-item sequential patterns. Early work \cite{zimdars2001using, rendle2010factorizing, wang2015learning} mostly focus on Markov chain \cite{norris1998markov} models. \cite{shani2005mdp} employs Markov decision processes in the recommender system to provide recommendations using sequential information. With the success of deep learning, researchers adopt (deep) neural network \cite{hidasi2015session,li2017neural,yuan2018simple,ma2019hierarchical} to model the sequential dynamics. Particularly, \cite{hidasi2015session} uses Gated Recurrent Units to encode previous behaviors into a hidden vector for the recommendation. Besides that, \cite{tang2018personalized} proposes a sequential model to learn sequential patterns using both horizontal and vertical convolutional filters. Recently, self-attention \cite{vaswani2017attention} attains promising results in various NLP tasks. \cite{kang2018self} firstly adopts the self-attention mechanism for sequential user behavior modeling, achieving state-of-the-art performance on the sequential recommendation.

Although the aforementioned methods achieve satisfactory results on sequential user behavior modeling task, they ignore the problem of long-tailed distribution, which may cause performance degradation for tail users. In this paper, we address this issue by learning transferrable parameters.

\subsection{Gradient Alignment}
The idea of gradient alignment has been well studied in various fields. Leap \cite{flennerhag2018transferring} utilizes gradient alignment to transfer knowledge across the learning process. \cite{riemer2018learning} attempts to solve the continual learning problem by considering a temporally symmetric trade-off between transfer and interference, which is implemented by encouraging gradient alignment across examples. More recently, \cite{zhang2019lookahead} presents the Lookahead optimization method, which improves learning stability and achieves faster convergence. In each step, the updating rule encourages the model parameters towards the aligned direction of gradients generated by different mini-batches. Different from previous works, we use gradient alignment to learn transferrable parameters in sequential user behavior modeling task.

\subsection{Adversarial Training}
Adversarial training is a well-studied problem \cite{szegedy2013intriguing,goodfellow2014explaining}, in which two or more models learn together by pursuing competing goals. A representative work of adversarial training is Generative Adversarial Nets \cite{goodfellow2014generative,salimans2016improved}, in which a discriminator and a generator compete with each other: the generator aims to generate samples similar to the real ones from random noise, and the discriminator aims to distinguish between the generated and the real samples. The networks are trained jointly using backpropagation on the prediction loss in a mini-max fashion: update generator to minimize the loss while also updating discriminator to maximize the loss. In this work,
we try to learn frequency-agnostic sequential embedding with the help of adversarial training. Specifically, we introduce a discriminator to differentiate sequential embeddings of head users and tail users while the prediction model aims to fool the discriminator to misclassify users and minimize the prediction loss simultaneously. In this way, the bias caused by the long-tail distribution of the number of user behaviors could be eliminated. A similar optimization approach has shown promising results in the neural word embedding literature \cite{gong2018frage}.

\section{PROPOSED METHODOLOGY}
\begin{figure*}[htb]
    \centering
    \includegraphics[width=\linewidth]{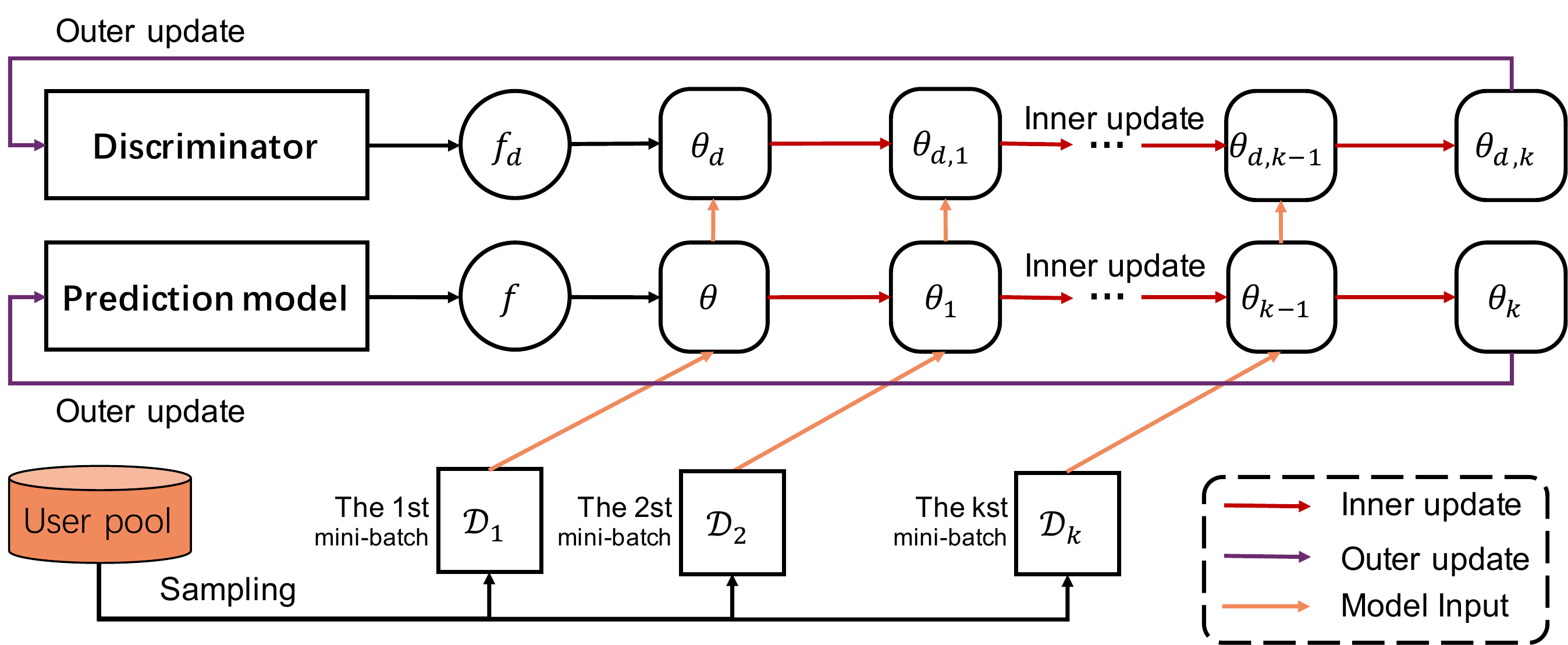}
    \caption{The structure of the proposed method. Each iteration consists of $k$ inner updates and one outer update. In each inner update phase (the red arrows), we select a user and sample a mini-batch of $L$-length subsequences from corresponding task for training. The parameters of the prediction model and the discriminator are updated with gradient descent. In outer update phase (the purple arrows), we update the prediction model and the discriminator in the direction of difference between the initial parameters and updated parameters.}
    \label{fig:structure}
\end{figure*}

In this section, we first give the notations and preliminaries of sequential user behavior modeling. Then we present how to learn transferrable parameters across users from optimization perspective and feature perspective, respectively. The structure of the proposed method is illustrated in Figure \ref{fig:structure}. 

\subsection{Problem Formulation}
Assume we have a set of users $\mathbf{U} = \{ u_1, u_2, \cdots, u_{|\mathbf{U}|}\}$ and a universe of items $\mathbf{I} = \{ i_1, i_2, \cdots, i_{|\mathbf{I}|}\}$. Each user $u$ is associated with a sequence of items sorted by time, which is represented as $\mathbf{S}^u = \{ S_1^u ,S_2^u , \cdots, S_t^u, \cdots, S_{|\mathbf{S}^u|}^u \}$, where $S_t^u \in \mathbf{I}$ denotes a user $u$ ever interacted with the item $S_t^u$ at time $t$. The objective is to seek a prediction model such that for a given prefix item subsequence $\mathbf{S}^{(u,t)} = \{ S_{t-L}^u ,S_{t-L+1}^u , \cdots, S_{t-1}^u \}$ including the most recent $L$ items before time $t$, it can generate a ranking score for all items:
\begin{equation}
    \mathbf{y}_u = f(\mathbf{S}^{(u,t)}; \theta).
\end{equation}
The prediction model $f$ is a composition of the sequence embedding function $\phi$ and a score function $h$, which can be written as $f(\cdot)=h(\phi(\cdot))$. We denote the parameters of prediction model $f$ as $\theta$. The binary cross entropy loss is often used as the optimization target \cite{tang2018personalized,kang2018self}:
\begin{equation}
    \mathcal{L} = \sum_{u} \sum_{j \in \mathbf{S}^u} [ - \log (\sigma(\mathbf{y}_{u,j})) - \sum_{k \notin \mathbf{S}^u} \log (1 - \sigma(\mathbf{y}_{u,k})) ],
\end{equation}
where $\sigma$ is the sigmoid function. Since the number of non-interactive items is relatively large, we follow the negative sampling strategy \cite{rendle2010factorizing,tang2018personalized} to choose the negative instances. The major notations in this paper are listed in table \ref{TBL:notations}.

\begin{table}[htb]
    \caption{List of notations.}
    \label{TBL:notations}
    \begin{tabular} {l l}
      \toprule
      Notation     & Meaning \\
      \midrule
      $\mathbf{U}$, $\mathbf{I}$    & the sets of users and items \\
      $\mathbf{S}^u$                & the item sequence of user $u$ \\
      $f, \theta$                   & the prediction model and parameters \\
      $f_d, \theta_d$               & the discriminator and parameters \\
      $\mathbf{y}_{u}$              & the prediction scores of user $u$ \\
      $\alpha, \beta$               & the learning rates of inner and outer update \\
      $k$                           & the inner update times \\
      $\lambda$                     & the adversarial hyper-parameter \\
      \bottomrule
    \end{tabular}
\end{table}

The problem of long tails makes existing sequential models perform poorly on tail users. To address this problem, we formulate each user as an individual task and then learn transferrable parameters across different tasks. For each user $u$, we extract every $L$ successive items as input and their next one item as the target from $\mathbf{S}^u$. Then we can get a corresponding training task $\mathcal{T}_u = \{ (\mathbf{S}^{(u,t)}, S_t^u) | t \in \{ L+1, L+2, ..., |\mathbf{S}^u|\} \}$. Given a set of training tasks $\{ \mathcal{T}_1, \mathcal{T}_2, \cdots,  \mathcal{T}_{|\mathbf{U}|} \}$, we aim to strike a balance between the performance of data-rich head users and data-poor tail users by exploiting transferrable model parameters. 

\subsection{Gradient Alignment}
\begin{figure}[htb]
    \centering
    \subfigure[transfer]{
		\includegraphics[width=0.22\textwidth]{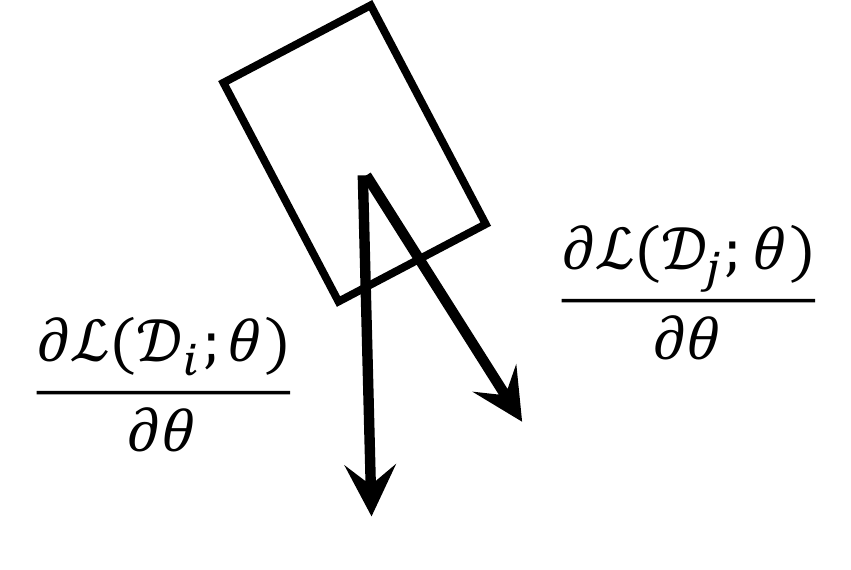}
    }
    \subfigure[interference]{
        \includegraphics[width=0.22\textwidth]{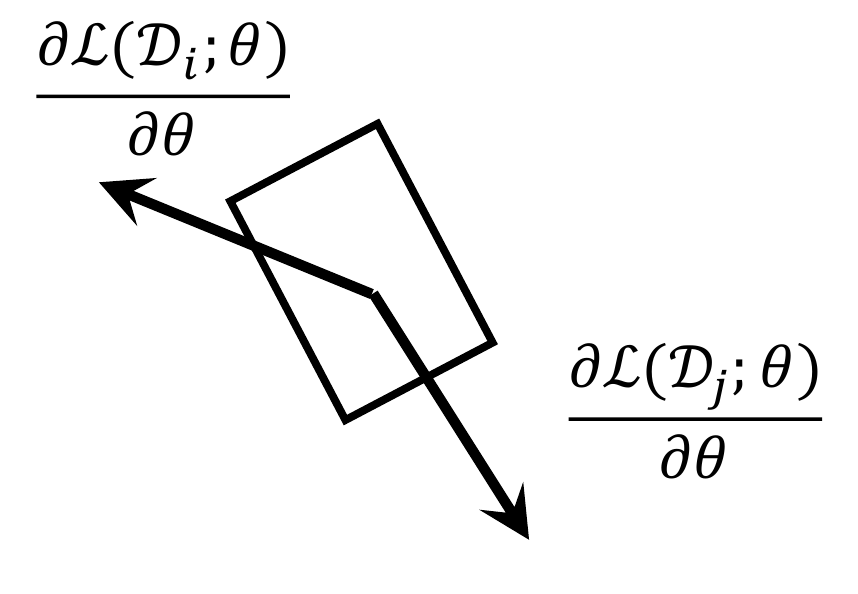}
    }
	\caption{(a) A depiction of transfer across users. (b) A depiction of interference across users.}
	\label{fig:transfer-interference-problem}
\end{figure}

The intuition behind our method is that when training with multiple tasks, some internal information are more transferrable across tasks while others can cause interference \cite{riemer2018learning}. Taking two users $i$ and $j$ as examples, the corresponding training tasks are $\mathcal{T}_i$ and $\mathcal{T}_j$, respectively. At each iteration, we sample $K$ subsequences from $\mathcal{T}_i$ and $\mathcal{T}_j$ separately for update. These two mini-batches are denoted as $\mathcal{D}_i$ and $\mathcal{D}_j$. Then, we can define operational measures of transfer and interference between these two distinct mini-batches. Formally, the concept of transfer is formulated as:
\begin{equation} \label{transfer}
    \frac{\partial \mathcal{L}(\mathcal{D}_i; \theta)}{\partial \theta} \cdot \frac{\partial \mathcal{L}(\mathcal{D}_j; \theta)}{\partial \theta} > 0,
\end{equation}
where $\cdot$ is the dot product operator. It indicates that solving the task of user $i$ will facilitate the learning process of the task of user $j$, and vice versa (Figure \ref{fig:transfer-interference-problem}(a)). In contrast, the concept of interference is formulated as:
\begin{equation} \label{interference}
    \frac{\partial \mathcal{L}(\mathcal{D}_i; \theta)}{\partial \theta} \cdot \frac{\partial \mathcal{L}(\mathcal{D}_j; \theta)}{\partial \theta} < 0.
\end{equation}
It implies that learning the task of user $i$ can impede the learning process of user $j$ and vice versa (Figure \ref{fig:transfer-interference-problem}(b)). The potential for transfer is maximized
when weight sharing is maximized while potential for interference is maximized when weight sharing is minimized. Since the tail users have limited data for training, encouraging the emergence of such transferrable parameters naturally transfer the knowledge from data-rich head users to data-poor tail users. Moreover, the transferrable parameters enjoy a better generalization capability, which facilitate the prediction of new users.


To maximize transfer and minimize interference, we incur an auxiliary loss to the objective, which can  bias the learning process to that direction. According to Eq. (\ref{transfer}) and (\ref{interference}), we evaluate the gradients of randomly sampled mini-batches from different users. By maximizing the inner product between the gradients of different mini-batches, we promote the model transferability by sharing parameters where gradient directions align. To this end, we can express our optimization objective as:
\begin{equation} \label{eq:integration}
\begin{split}
    \theta = & arg\min_{\theta} \mathbb{E}_{\mathcal{D}_i \& \mathcal{D}_j} \\
    & \left[ \mathcal{L}(\mathcal{D}_i; \theta) + \mathcal{L}(\mathcal{D}_j; \theta) - \alpha \frac{\partial \mathcal{L}(\mathcal{D}_i; \theta)}{\partial \theta} \cdot \frac{\partial \mathcal{L}(\mathcal{D}_j; \theta)}{\partial \theta} \right].
\end{split}
\end{equation}
Overall, the first two terms focus on the minimum of the expected loss over tasks, while the last term enables knowledge transfer across users by maximizing the inner product between different gradients. 

However, directly optimizing the above objective requires to explicitly compute second derivatives, which is quite expensive. Motivated by \cite{nichol2018first}, we approximate the objective in (\ref{eq:integration}) with a first order Taylor expansion to reduce computational overhead. Specifically, the training process is divided into two phases: inner update and outer update. In inner update phase, we perform $k$ gradient updates with learning rate $\alpha$. At each time, we randomly select one user and sample a mini-batch from corresponding task for update:
\begin{equation}
    \theta_k = \theta_{k-1} + \alpha \frac{\partial \mathcal{L}(\mathcal{D}_k;\theta_{k-1})}{\partial \theta_{k-1}}.
\end{equation}
These inner updates start with parameters $\theta$ and result in parameters $\theta_k$. Then we enter outer update phase: updating the model parameters $\theta$ in the direction $\theta - \theta_k$:
\begin{equation}
    \theta = \theta - \beta *(\theta - \theta_k),
\end{equation}
where $\beta$ is the learning rate. According to taylor's theorem, the above update process is approximately equivalent to optimize the following objective:
\begin{equation}
\label{eq:approximation}
\begin{split}
    \theta = & arg\min_{\theta} \mathbb{E}_{\mathcal{D}_1,\cdots, \mathcal{D}_k}\\
    &\left[ \sum_{i=1}^{k} \left[\mathcal{L}(\mathcal{D}_i; \theta) - \frac{1}{2} \sum_{j=1}^{i-1} \alpha \frac{\partial \mathcal{L}(\mathcal{D}_i; \theta)}{\partial \theta} \cdot \frac{\partial \mathcal{L}(\mathcal{D}_j; \theta)}{\partial \theta} \right] \right],
\end{split}
\end{equation}
where $\mathcal{D}_1, \cdots, \mathcal{D}_s$ are mini-batches sampled from different tasks. Compared to the original objective in (\ref{eq:integration}), Eq. (\ref{eq:approximation}) also contains the transferability-promoting terms. The difference lies in the importance of different users. Eq. (\ref{eq:approximation}) assigns high importance to the early chosen user in the inner update. As we randomly choose user at each inner update, Eq. (\ref{eq:approximation}) does not bring any bias towards users.
	
\subsection{Adversarial Training}
Although gradient alignment could encourage knowledge transfer across users, the prediction model, especially for the sequence embedding function, will be largely dominated by the head users which have more data available for training. More severely, the sequential embeddings of head users and tail users tend to lie in different spaces. It will consequently limit the performance of prediction using the embeddings. To address this issue, we develop an adversarial training method \cite{goodfellow2014generative} to mix head users and tail users into a common embedding space and thus produce frequency-agnostic sequential embedding.  Specifically, we adopt a discriminator to categorize users into two classes: head or tail. We hope that the discriminator optimizes its parameters to maximize its classification accuracy, while the prediction model is optimized towards a low training loss as well as fooling the discriminator to misclassify head users and tail users. When the whole optimization process converges, the discriminator cannot well classify head users and tail users. In this way, the bias caused by the large amount of data from head users could be reduced, thus facilitating knowledge transfer from the head to the tail.

To begin with, we divide users into two parts based on the number of historical behaviors and use $R \in \{0,1\}$ to indicate this property: $R = 1$ for head users; $R = 0$ for tail users. Let $f_{d}$ and $\theta_{d}$ denote the discriminator and its parameters respectively, then the loss of the discriminator $\mathcal{L}_{d}$ can be defined as:
\begin{equation}
\begin{split}
    \mathcal{L}_{d}(\mathcal{D};\theta,\theta_d) = & R \log f_{d} (\phi(\mathcal{D}; \theta);\theta_{d}) \\
    & + (1 - R) \log (1 - f_{d} (\phi(\mathcal{D}; \theta);\theta_{d}).
\end{split}
\end{equation}
Following the principle of adversarial training, we develop a minimax objective to train the prediction model and the discriminator as below:
\begin{equation} \label{MinMaxLoss}
    \min_{\theta,\theta_d} \max_{\theta_d} \mathcal{L}(\mathcal{D};\theta) - \lambda \mathcal{L}_d(\mathcal{D};\theta,\theta_d),
\end{equation}
where $\lambda$ is a hyper-parameter to trade off the two loss terms.

Now we summarize the training process of our method in Algorithm \ref{alg:TP}. In each iteration, we perform $k$ inner updates and one outer update. the proposed method requires $k \ge 2$, where the update depends on the higher-order derivatives of the loss function. Otherwise, if we set $k$ to be $1$, it will degrade to a classic joint optimization over all tasks. It is worth mentioning that our method is model agnostic and can be applied to any existing models for sequential recommendation, such as GRU4REC \cite{hidasi2015session}, CASER \cite{tang2018personalized} and SASR \cite{kang2018self}, which are used for our empirical study.

\begin{algorithm}
\caption{Proposed Algorithm}
\label{alg:TP}
\begin{algorithmic}
\STATE{\textbf{Input:} $\mathbf{U}$: the set of existing users}
\STATE{\textbf{Input:} $\mathcal{T}$: the set of training tasks}
\STATE{\textbf{Input:} $\alpha, \beta$: the learning rates of inner and outer update}
\STATE{\textbf{Input:} $k$: the inner update times}
\STATE{\textbf{Input:} $\lambda$: the adversarial hyper-parameter}
\STATE{Randomly initialize the prediction model parameters $\theta$ and the discriminator parameters $\theta_d$}
\FOR {$iter = 1, 2, \dots$}
    \STATE{$\Theta \gets \theta$, $\Theta_d \gets \theta_d$};
    \FORALL {$i = 1, 2, ..., k$}
        \STATE{Select a user $u$ and corresponding task $\mathcal{T}$};
        \STATE{Sample a mini-batch $\mathcal{D}$ from $\mathcal{T}$};
        \STATE{Update $\theta, \theta_d$ by SGD according to Eq. (\ref{MinMaxLoss}) with learning rate $\alpha$};
    \ENDFOR
    \STATE{Update the prediction model: $\theta \gets \Theta + \beta (\Theta-\theta)$};
    \STATE{Update the discriminator: $\theta_d \gets \Theta_d + \beta (\Theta_d-\theta_d)$};
\ENDFOR
\end{algorithmic}
\end{algorithm}

\section{EXPERIMENTS}
In this section, we conduct experiments on real-world datasets to evaluate the proposed method. We aim to answer the following research questions:

\begin{enumerate} [\textbf{RQ} 1]
    \item How effective is the proposed method compared with the state-of-the-art competitors?
    \item Does the proposed method actually improve the performance of tail users?
    \item How do the gradient alignment and adversarial training affect the performance of the proposed method?
\end{enumerate}

\subsection{Experimental Setting}
\subsubsection{Dataset}
We conduct experiments on the following publicly accessible datasets: Amazon\footnote{http://jmcauley.ucsd.edu/data/amazon/}, MovieLens\footnote{http://grouplens.org/datasets/movielens/} and MovieTweetings\footnote{https://github.com/sidooms/MovieTweetings}.

\begin{table*}[htb]
    \caption{Statistics of the evaluation datasets.}
    \label{TBL:Dataset}
    \centering
        \begin{tabular} {l | r r r r r r}
            \hline
            Dataset         & \#Users & \#Items    & \#Interactions   & \#Records/user  & \#Records/item  & \#Density \\
            \hline
            Music           & 2,831   & 13,410     & 63,054           & 22.27           & 4.70           & 0.163 \% \\
            Game            & 7,519   & 19,977     & 145,520          & 19.35           & 7.28           & 0.097 \% \\
            ML1M            & 6,040   & 3,706      & 1,000,209        & 165.57          & 269.89         & 4.468 \% \\
            MovieTweetings  & 12,605  & 10,910     & 654,658          & 51.94           & 60.01          & 0.476 \% \\
            \hline
        \end{tabular}
\end{table*}

\textbf{Amazon.} This dataset contains a series of product purchase histories crawled from the Amazon website. Top-level product categories are used to split the dataset into separate subsets. Here, we conduct experiments on the Digital Music (\textbf{Music}) and Video Games (\textbf{Game}) category.

\textbf{MovieLens.} This movie rating dataset is widely used in recommendation tasks. In our experiment, we use the version that includes 1 million user ratings: MovieLens 1M (\textbf{ML1M}).

\textbf{MovieTweetings.} This is a dataset consisting of ratings on movies that are contained in well-structured tweets on Twitter. 

For dataset preprocessing, we follow the common practice in previous works \cite{he2016fusing,zhang2019next,yu2019multi}. For all datasets, we convert explicit ratings into implicit binary feedbacks. After that, we group interactions by user ID, and construct each user's sequence by sorting according to the timestamps. In order to reduce the impact of noise data, we filter out inactive items with fewer than 5 related records, and then remove users with less than 10 feedbacks. Detail statistics of the datasets are summarized in table \ref{TBL:Dataset}. Finally, each dataset is randomly divided into two parts according to users: 80\% of the users as existing users, and the remaining 20\% as new users for the analysis of cold start problem. For existing users, we use the most recent item in the interaction sequence of each user for evaluation, and the rest items for training.

\subsubsection{Evaluation Metrics}
The following metrics are used to evaluate the quality of recommendation, which are also widely used in previous works \cite{kang2018self,yu2019multi,zhang2019next}.

\textbf{HR@N} (Hit Ratio) is widely used as a measure of predictive accuracy. It represents the proportion of the desired item amongst the top-N items in all test cases. It is computed as:

\begin{equation}
  {\rm HR}@N = \frac{1}{|\mathbf{U}|} \sum_{u \in \mathbf{U}} \mathbb{I} (R_{u, g_u} \leq N),
\end{equation}
where $\mathbb{I}$ is an indicator function. $g_u$ is the test item of user $u$, and $R_{u,g_u}$ is the rank of $g_u$ generated by the model.

\textbf{NDCG@N} (Normalized Discounted Cumulative Gain) records the position of the hit by assigning larger scores on higher ranks. It is computed as:
\begin{equation}
  {\rm NDCG}@N = \frac{1}{|\mathbf{U}|} \sum_{u \in \mathbf{U}} \frac{\mathbb{I} (R_{u, g_u} \leq N)}{log_2 {(R_{u, g_u} + 1)}}.
\end{equation}

Basically, the higher these metrics, the better the performance of the model. To make results more stable, we repeate each experiment ten times for each metric and compute the average results.

\subsubsection{Baselines}
To demonstrate the effectiveness of the proposed method, we compare to the following sequential user behavior modeling methods.

\textbf{- GRU4REC.} \cite{hidasi2015session} This is a representative deep learning based method which adopts RNN to model users' sequential behaviors.

\textbf{- CASER.} \cite{tang2018personalized} This CNN based method attempts to capture sequential patterns on both individual-level and union-level with the help of horizontal and vertical convolutional filters.

\textbf{- SASR.} \cite{kang2018self} This is a self-attention based sequential model which adopts an attention mechanism to identify relevant items for predicting the next item.

The above three baselines are representative sequential user behavior modeling methods based on RNN, CNN and self-attention, respectively. Since our framework is model agnostic, we use it to train the above three baselines.
For fair comparisons, models trained by our method shares the same model architecture as the baselines. For simplicity, we use ``-TP'' denote models trained by our method compared with baselines. In the experiments, we adopt grid search to select the best parameters for each model. The embedding size is turned from $\{10, 15, \cdots, 50\}$, the regularization parameter and learning rate are selected from $\{1e^{-4}, 1e^{-3}, \cdots, 1\}$. All other hyper-parameters and initialization strategies are those suggested by the methods' authors. For our method, we adopt SGD optimizer and fix the update times $k=2$ in inner update phase. For outer update phase, we employ adam optimizer, whose learning rate $\beta$ is adam's suggested setting $0.001$. For adversarial training, we simply set 20\% users with the most interactions as head users and the rest as tail users, which is the same as the pareto principle \cite{sanders1987pareto}. The adversarial hyper-parameter $\lambda$ is tuned from $\{0, 0.01, 0.1, 1, 10\}$.

\begin{table*}[htb]
    \caption{Performance comparison of all methods in terms of HR@10 and NDCG@10. The best results are boldfaced.}
    \label{TBL:Result}
    \begin{tabular}{l|l|c|c|c|c|c|c|c|c}
    \hline
    \multirow{2}{*}{Type} & \multirow{2}{*}{Method} & \multicolumn{2}{c|}{Music} & \multicolumn{2}{c|}{Game} & \multicolumn{2}{c|}{ML1M} & \multicolumn{2}{c}{MovieTweetings} \\ \cline{3-10} 
     &  & HR@10 & NDCG@10 & HR@10 & NDCG@10 & HR@10 & NDCG@10 & HR@10 & NDCG@10 \\ \hline
    \multirow{6}{*}{Existing users} 
     & GRU4REC & 0.0238 & 0.0125 & 0.0324 & 0.0167  & 0.2036 & 0.1090 & 0.1482 & 0.0767\\ 
     & GRU4REC-TP & \textbf{0.0287} & \textbf{0.0139} & \textbf{0.0359} & \textbf{0.0189} & \textbf{0.2194} & \textbf{0.1196} & \textbf{0.1586} & \textbf{0.0850} \\ \cline{2-10} 
     & CASER & 0.0499 & 0.0254 & 0.0598 & 0.0319 & 0.2314 & 0.1278 & 0.1502 & 0.0786 \\
     & CASER-TP & \textbf{0.0556} & \textbf{0.0314} & \textbf{0.0638} & \textbf{0.0330} & \textbf{0.2469} & \textbf{0.1335} & \textbf{0.1596} & \textbf{0.0870} \\ \cline{2-10} 
     & SASR & 0.0530 & 0.0295 & 0.0607 & 0.0298 & 0.2210 & 0.1133 & 0.1516 & 0.0787 \\ 
     & SASR-TP & \textbf{0.0605} & \textbf{0.0320} & \textbf{0.0640} & \textbf{0.0321} & \textbf{0.2411} & \textbf{0.1256} & \textbf{0.1629} & \textbf{0.0876} \\ \hline \hline
    \multirow{6}{*}{New users} 
     & GRU4REC & 0.0194 & 0.0102 & 0.0266 & 0.0135 & 0.1945 & 0.0999 & 0.1444 & 0.0771 \\ 
     & GRU4REC-TP & \textbf{0.0247} & \textbf{0.0111} & \textbf{0.0306} & \textbf{0.0156} & \textbf{0.2094} & \textbf{0.1058} & \textbf{0.1535} & \textbf{0.0823} \\ \cline{2-10} 
     & CASER & 0.0442 & 0.0230 & 0.0532 & 0.0259 & 0.2119 & 0.1116 & 0.1420 & 0.0771 \\ 
     & CASER-TP & \textbf{0.0512} & \textbf{0.0291} & \textbf{0.0552} & \textbf{0.0280} & \textbf{0.2293} & \textbf{0.1185} & \textbf{0.1523} & \textbf{0.0843} \\ \cline{2-10} 
     & SASR & 0.0495 & 0.0242 & 0.0532 & 0.0259 & 0.2127 & 0.1083 & 0.1440 & 0.0750 \\ 
     & SASR-TP & \textbf{0.0548} & \textbf{0.0295} & \textbf{0.0566} & \textbf{0.0282} & \textbf{0.2227} & \textbf{0.1147} & \textbf{0.1599} & \textbf{0.0860} \\ \hline
    \end{tabular}
\end{table*}

\begin{figure*}
	\centering
    \subfigure[Recommendation for existing users]{
		\includegraphics[width=\linewidth]{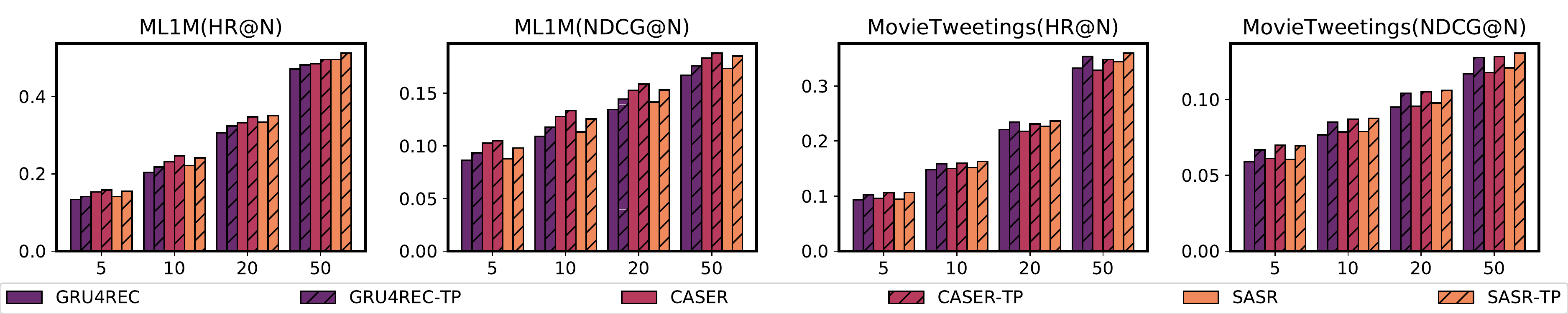}
    }
    \subfigure[Recommendation for new users]{
        \includegraphics[width=\linewidth]{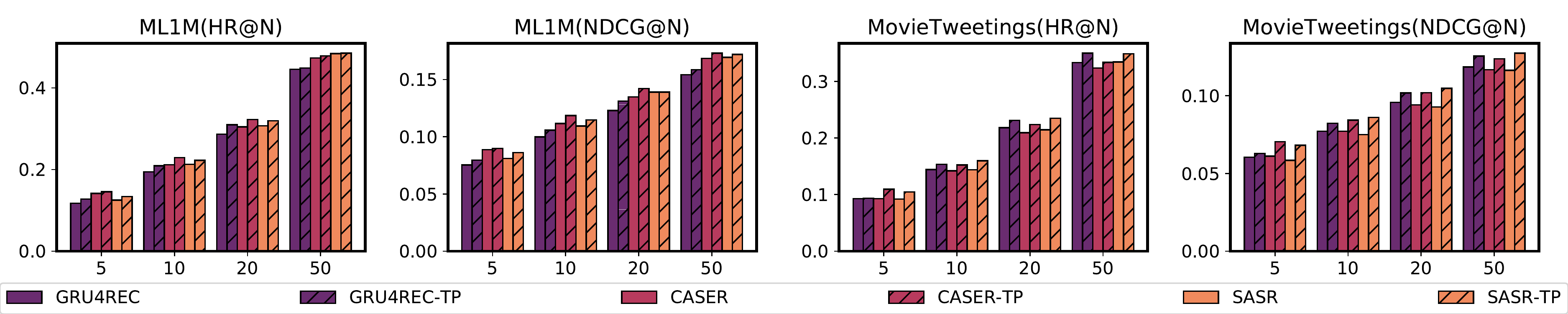}
    }
	\caption{Top-N recommendation evaluation with different values of $N$ on HR and NDCG.}
	\label{fig:metric}
\end{figure*}

\subsection{Performance Comparison}
To answer \textbf{RQ1}, we evaluate the recommendation performance of the proposed method and all baselines. Results are shown in Table \ref{TBL:Result} and we can draw following observations.

\textbf{Observations about our framework.}
First, for existing users, models trained using the proposed method consistently achieve better performance than their conventional counterparts, showing the benefits of learning transferrable parameters. Our method encourages knowledge transfer across users, leading to a general model. Such a model will not be biased to either head or tail users but achieves good performance on both of them. Second, all methods perform worse on new users than existing users, which indicates that these sequential models cannot handle user cold-start problem well. Generally speaking, sequential models can make recommendations based on previous interactions without relying on the user profile. However, sequential patterns learned from existing users may not suitable for new users, which leads to performance degradation for new users. Third, models trained by our method outperform their conventional counterparts when facing new users. This result indicates that the proposed method can effectively enhance the sequential model's ability to deal with the cold start problem of new users. The reason is that our method focuses on those more diverse and noisy tail users, which is helpful for learning a more robust model that can perform well on new users. Finally, besides the above evaluation of different methods, Figure \ref{fig:metric} shows models trained by our method are stably superior to their conventional counterparts with different lengths of the recommendation list. 

\textbf{Other observations.}
First, all methods achieve better results on MovieTweetings and ML1M datasets than Music and Game datasets. The major reason is that Music and Game datasets are more sparse than MovieTweetings and ML1M datasets, and the data sparsity declines the recommendation performance. Second, SASR outperforms than GRU4REC and CASER in most cases. The main reason is that SASR adopts the self-attention mechanism to attend items adaptively that would better reflect the user's preference. Finally, the best performing models are not consistent on different datasets, which suggests that we should choose the appropriate model according to the actual situation. It should be emphasized that our method is model agnostic and can be well adapted to various well-established sequential models.

\begin{figure*}
    \centering
    \subfigure[Recommendation for existing users]{  
        \includegraphics[width=\linewidth]{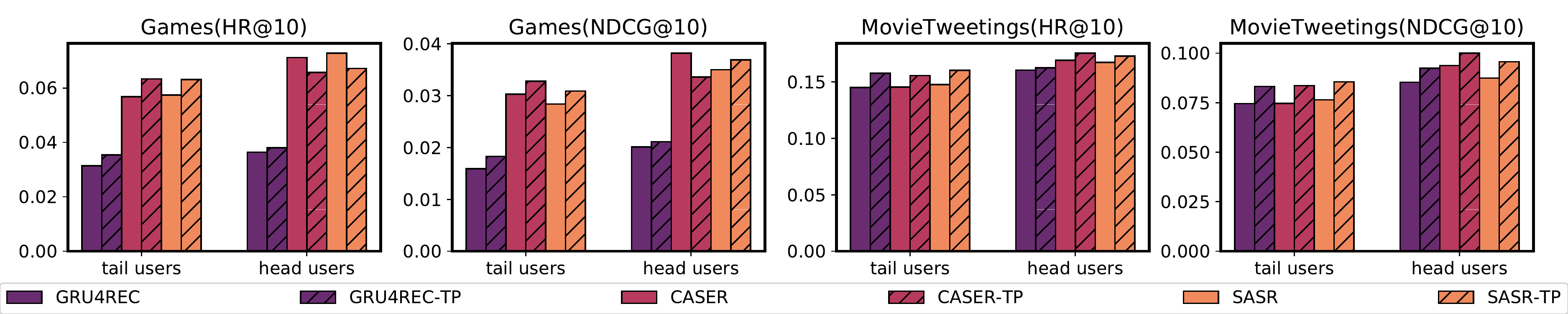}
    }
    \subfigure[Recommendation for new users]{
        \includegraphics[width=\linewidth]{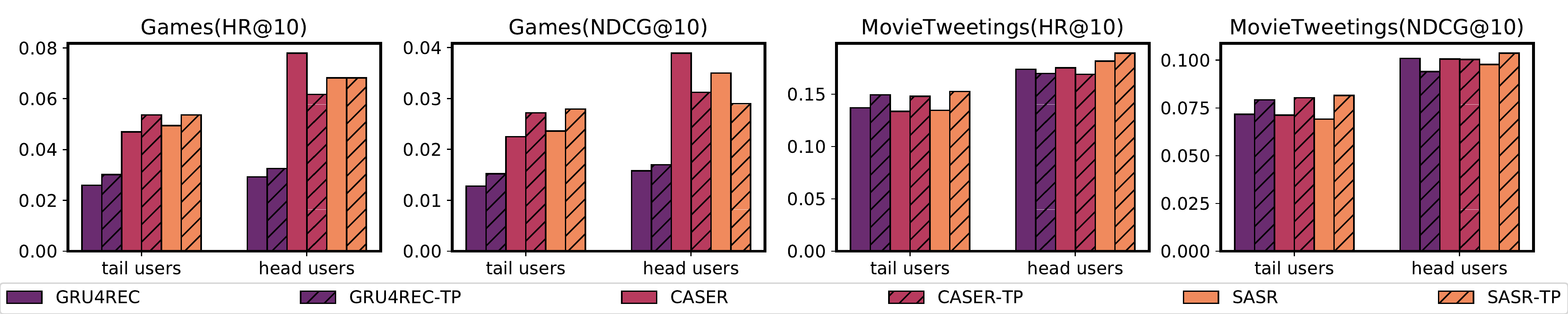}
    }
    \caption{Recommendation evaluation on head users and tail users.}
    \label{fig:user-frequency}
\end{figure*}

\begin{figure*}
    \centering
    \includegraphics[width=\linewidth]{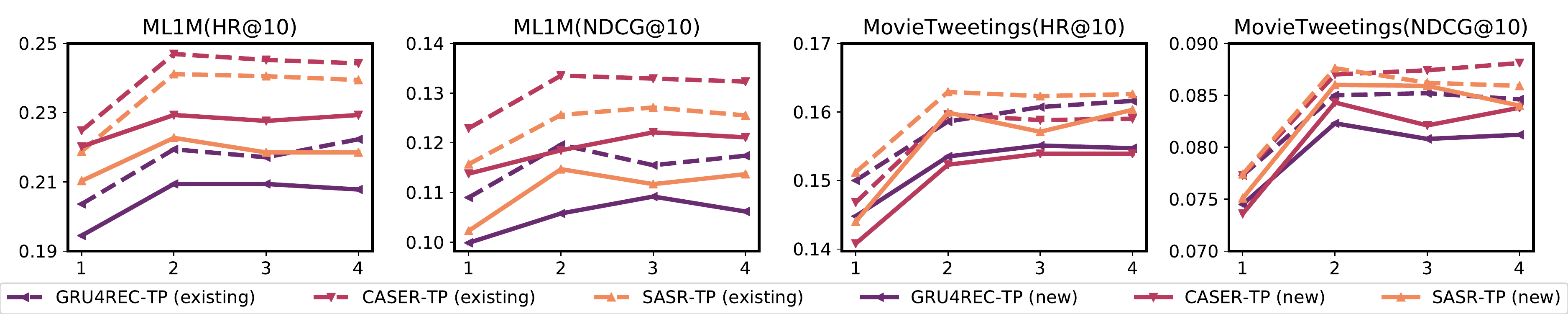}
    \caption{Impact of inner update times $k$.}
    \label{fig:inner}
\end{figure*}

\begin{figure*}
    \centering
    \includegraphics[width=\linewidth]{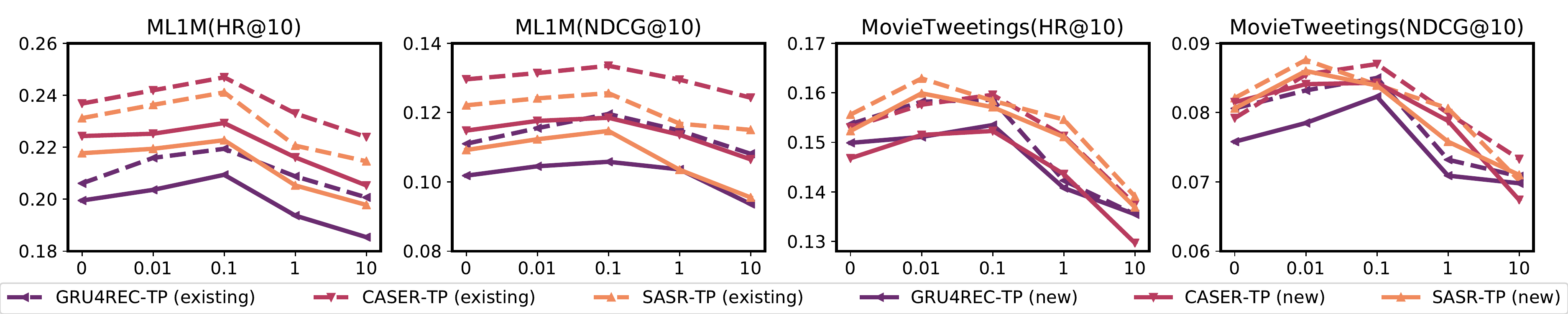}
    \caption{Impact of adversarial parameter $\lambda$.}
    \label{fig:adversarial-parameter}
\end{figure*}

\subsection{Performance on Head Users and Tail Users}
To answer RQ2, we divide the evaluation by the number of user's historical interactions. Specifically, for both existing and new users, we treat 20\% of the users with the most historical interactions as head users and denote the rest as tail users, which is the same as the training phase. The results are shown in Figure \ref{fig:user-frequency}. 

First of all, we can notice that all methods perform better on head users than on tail users. This result confirms our conjecture that head users have more data available for training than tail users, resulting in better performance. Second, for both existing and new users, models trained by our method achieve significant improvements on tail users than their conventional counterparts. On the other hand, the performance improvement of head users brought by our method is relatively small compared to tail users. Even in some cases, there is a certain decrease in the performance of head users (i.e., CASER-TP compared to CASER on both Game and MovieTweetings datasets for new users). In other words, models trained using our method will pay more attention to tail users, which helps achieve a balanced performance instead of biased towards head users. This result demonstrates that the proposed method can effectively transfer knowledge from head users to tail users, thereby achieving a general model that performs well on both head users and tail users. Finally, by comparing Game and MovieTweetings datasets, we find that the performance gap between head users and tail users on Game dataset is larger than MovieTweetings dataset. Moreover, models trained using our method can achieve a more balanced performance on Game dataset than MovieTweetings dataset, and the performance improvement of tail users is more significant. The reason could be that Game dataset is more sparse and the information gap between head users and tail users is larger than MovieTweetings dataset. In contrast, MovieTweetings dataset is relatively dense. Tail users have more data, enough to achieve good performance. Therefore, the performance improvement brought by learning transferrable parameters of our method is weaker on MovieTweetings dataset than Game dataset.
 
\subsection{Impact of Gradient Alignment and Adversarial Training}
For the proposed method, there are two key parameters related to gradient alignment and adversarial training. The first is the inner update times $k$. Figure \ref{fig:inner} shows the performance of our method for varying inner update times $k$ on ML1M and MovieTweetings datasets. The second is the adversarial hyper-parameter $\lambda$. Figure \ref{fig:adversarial-parameter} shows the performance of our method for varying adversarial hyper-parameter $\lambda$ on ML1M and MovieTweetings datasets.

First, by examining the influence of the inner update times $k$, we found that the performance of our method is worst when $k = 1$. This is reasonable since it only optimizes the expected loss over all users, without considering the gradient alignment between different users, which can learn transferrable parameters from an optimization perspective. By further investigating the performance when $k \ge 2$, we found that there are only slight differences when increasing the inner update times compared to the significant improvement as $k$ from 1 to 2. One possible explanation is that knowledge transfer across users can be achieved for any value of $k \ge 2$. Increasing the value of $k$, which is greater than 2 cannot significantly change the effect of knowledge transfer.

Second, by examining the influence of $\lambda$, we found that better performance is achieved by balancing the impact of prediction loss and adversarial loss, while either a large or small value of $\lambda$ will adversely degrade the performance. Presumably, this is because a too large value of $\lambda$ means the prediction model spends too much effort to learn frequency-agnostic sequential embedding, resulting in insufficient sequential modeling problem. On the contrary, a too small value of $\lambda$ cannot effectively transfer knowledge from head users to tail users, resulting in a biased model, then reducing the performance of our method.

\section{CONCLUSION}
In this paper, we propose to solve the long-tailed distribution problem in sequential user behavior modeling by learning transferrable parameters. Specifically, we propose a gradient alignment optimizer to encourage knowledge transfer from the optimization perspective. Moreover, we introduce an adversarial training method to learn frequency-agnostic sequential embedding, which facilitates knowledge transfer from the feature perspective. Experiments on real-world datasets demonstrate the effectiveness of the proposed method, by comparing with the state-of-the-art baselines.

\section*{Acknowledgments}
This research is supported by the National Research Foundation, Singapore under its AI Singapore Programme (AISG Award No: AISG-RP-2018-001). Any opinions, findings and conclusions or recommendations expressed in this material are those of the author(s) and do not reflect the views of National Research Foundation, Singapore.

\bibliographystyle{ACM-Reference-Format}
\balance
\bibliography{sample-base}

\end{document}